\documentclass[10pt,twocolumn,letterpaper]{article}

\usepackage{cvpr}
\usepackage{times}
\usepackage{epsfig}
\usepackage{graphicx}
\usepackage{amsmath}
\usepackage{amssymb}

\usepackage{comment}
\renewcommand{\thefootnote}{}
\usepackage{amssymb}
\usepackage{pifont}
\newcommand{\cmark}{\ding{51}}%
\newcommand{\xmark}{\ding{55}}%

\makeatletter
\newcommand{\printfnsymbol}[1]{%
  \textsuperscript{\@fnsymbol{#1}}%
}
\makeatother

\usepackage[pagebackref=true,breaklinks=true,letterpaper=true,colorlinks,bookmarks=false]{hyperref}

\cvprfinalcopy 

\def\cvprPaperID{928} 

\ifcvprfinal\fi
\begin{document}
\bibliographystyle{unsrt}

\title{Few-Shot Object Detection with Attention-RPN and Multi-Relation Detector} 

\author{Qi Fan\thanks{Both authors contributed equally.} \ 
\\ HKUST\\
{\tt\small qfanaa@cse.ust.hk}
\and
Wei Zhuo\printfnsymbol{1}\\
Tencent\\
{\tt\small wei.zhuowx@gmail.com}
\and
Chi-Keung Tang\\
HKUST\\
{\tt\small cktang@cse.ust.hk}
\and
Yu-Wing Tai\\
Tencent\\
{\tt\small yuwingtai@tencent.com}
}

\maketitle

\begin{abstract}
    Conventional methods for object detection typically require a substantial amount of training data and preparing such high-quality training data is very labor-intensive. In this paper, we propose a novel few-shot object detection network that aims at detecting objects of unseen categories with only a few annotated examples. Central to our method are our Attention-RPN, Multi-Relation Detector and Contrastive Training strategy, which exploit the similarity between the few shot support set and query set to detect novel objects while suppressing false detection in the background. To train our network, we contribute a new dataset that contains 1000 categories of various objects with high-quality annotations. To the best of our knowledge, this is one of the first datasets specifically designed for few-shot object detection. Once our few-shot network is trained, it can detect objects of unseen categories without further training or fine-tuning. Our method is general and has a wide range of potential applications. We produce a new state-of-the-art performance on different datasets in the few-shot setting.  The dataset link is \href{https://github.com/fanq15/Few-Shot-Object-Detection-Dataset}{https://github.com/fanq15/Few-Shot-Object-Detection-Dataset}. \footnote{This research is supported in part by Tencent and the Research Grant Council of the Hong Kong SAR under grant no. 1620818.}
\end{abstract}

\section{Introduction}

\renewcommand{\thefootnote}{1}

Existing object detection methods typically rely heavily on a huge amount of annotated data and require long training time. 
This has motivated the recent development of few-shot object detection. Few-shot learning is challenging given large variance of illumination, shape, texture, etc, in real-world objects.  While significant research and progress have been made~\cite{snell2017prototypical, Sachin2017, santoro2016meta,vinyals2016matching, Finn2017ModelAgnosticMF,cai2018memory,gidaris2018dynamic,yang2018learning}, all of these methods focus on image classification rarely tapping into the problem of few-shot object detection, most probably because transferring from few-shot classification to few-shot object detection is a non-trivial task. 

Central to object detection given only a few shots is how to localize an unseen object in a cluttered background, which in hindsight is a general problem of object localization from a few annotated examples in novel categories. Potential bounding boxes can easily miss unseen objects, or else many false detections in the background can be produced. We believe this is caused by the inappropriate low scores of good bounding boxes output from a region proposal network (RPN) making a novel object hard to be detected. This makes the few-shot object detection intrinsically different from few-shot classification. Recent works for few-shot object detection~\cite{Chen2018LSTDAL, kang2018few, karlinsky2019repmet, yan2019metarcnn} on the other hand all require fine-tuning and thus cannot be directly applied on novel categories.

\begin{figure}[!t]
\centering
\includegraphics[width=7.5cm]{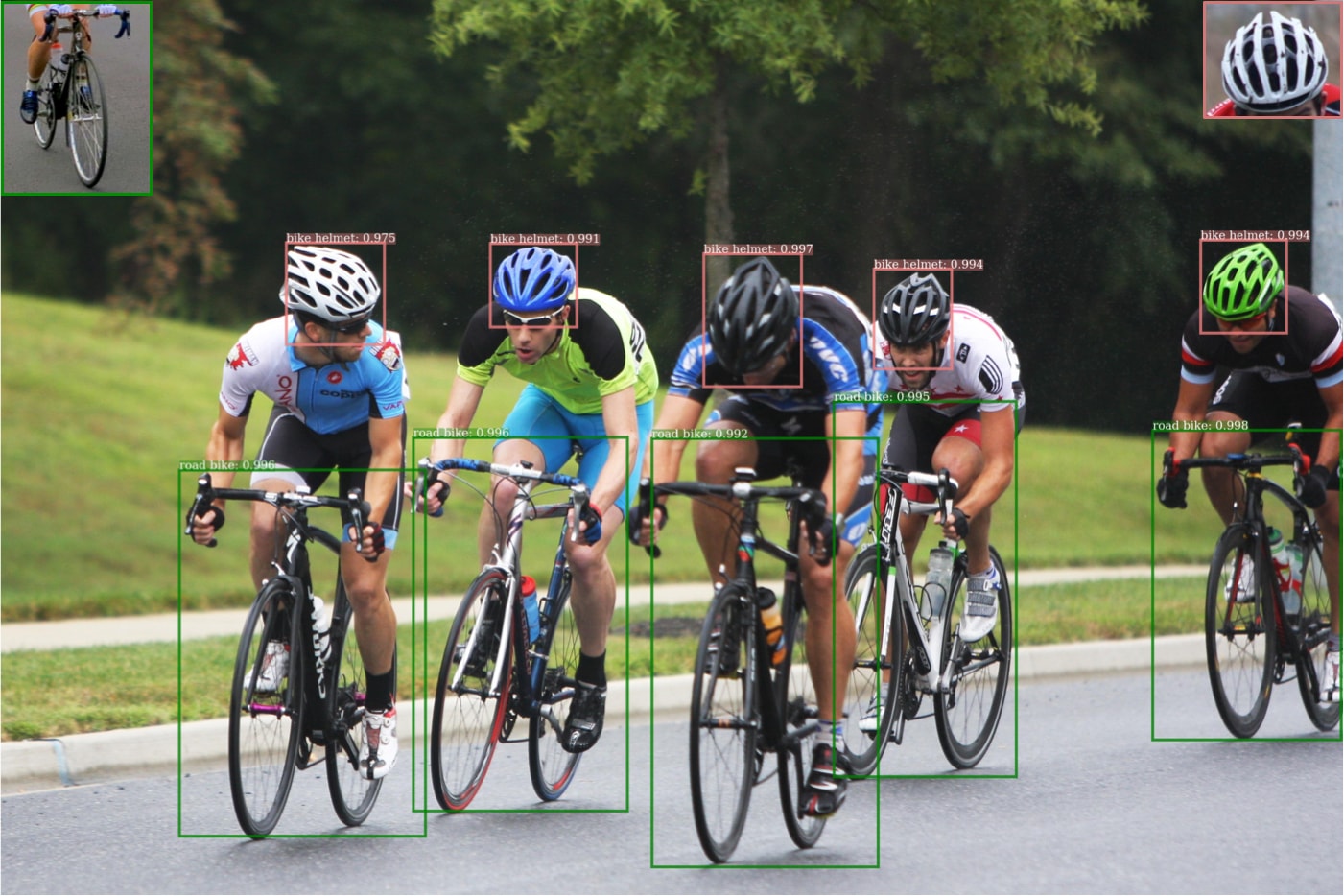}
\caption{Given different objects as supports (top corners above), our approach can detect all objects in the same categories in the given query image.}
\label{fig:intro}
\vspace{-0.6cm}
\end{figure}

In this paper, we address the problem of few-shot object detection: given a few support images of novel target object, our goal is to detect all foreground objects in the test set that belong to the target object category, as shown in Fig.~\ref{fig:intro}. To this end, we propose two main contributions:

First, we propose a general few-shot object detection model that can be applied to detect novel objects without re-training and fine-tuning. With our carefully designed contrastive training strategy, attention module on RPN and detector, our method exploits matching relationship between object pairs in a weight-shared network at multiple network stages.
This enables our model to perform online detection on objects of novel categories requiring {\em no} fine-training or further network adaptation.  Experiments show that our model can benefit from the attention module at the early stage where the proposal quality is significantly enhanced, and from the multi-relation detector module at the later stage which suppresses and filters out false detection in the confusing background. Our model achieves new state-of-the-art performance on the ImageNet Detection dataset and MS COCO dataset in the few-shot setting.

The second contribution consists of a large well-annotated dataset with 1000 categories with only a few examples for each category. Overall, our method achieves significantly better performance by utilizing this dataset than existing large-scale datasets, \eg COCO~\cite{lin2014microsoft}. To the best of our knowledge, this is one of the first few-shot object detection datasets with an unprecedented number of object categories (1000). Using this dataset, our model achieves better performance on different datasets even without any fine-tuning.

\section{Related Works}
\noindent {\bf General Object Detection.} Object detection is a classical problem in computer vision. In early years, object detection was usually formulated as a sliding window classification problem using handcrafted features~\cite{dalal2005histograms, felzenszwalb2010object, vioda2001rapid}. With the rise of deep learning~\cite{NIPS2012_4824}, CNN-based methods have become the dominant object detection solution.  Most of the methods can be further divided into two general approaches: proposal-free detectors and proposal-based detectors. The first line of work follows a one-stage training strategy and does not explicitly generate proposal boxes~\cite{redmon2016you, redmon2017yolo9000, liu2016ssd, lin2017focal, liu2018receptive}. On the other hand, the second line, pioneered by R-CNN~\cite{girshick2014rich}, first extracts class-agnostic region proposals of the potential objects from a given image. These boxes are then further refined and classified into different categories by a specific module~\cite{girshick2015fast, ren2015faster, he2017mask, singh2018sniper}. An advantage of this strategy is that it can filter out many negative locations by the RPN module which facilitates the detector task next. For this sake, RPN-based methods usually perform better than proposal-free methods with state-of-the-art results~\cite{singh2018sniper} for the detection task. 
The methods mentioned above, however, work in an intensive supervision manner and are hard to extend to novel categories with only several examples. 

\noindent {\bf Few-shot learning.}  Few-shot learning in a classical setting~\cite{thrun1996learning} is challenging for traditional machine learning algorithms to learn from just a few training examples. Earlier works attempted to learn a general prior~\cite{fei2006one, lake2011one, lake2013one, lake2015human, wong2015one}, such as hand-designed strokes or parts which can be shared across categories.  Some works~\cite{snell2017prototypical, oreshkin2018tadam, triantafillou2017few, hariharan2017low} focus on metric learning in manually designing a distance formulation among different categories. A more recent trend is to design a general agent/strategy that can guide supervised learning within each task; by accumulating knowledge the network can capture the structure variety across different tasks. This research direction is named meta-learning in general~\cite{Sachin2017, Finn2017ModelAgnosticMF, koch2015siamese, munkhdalai2017meta, Munkhdalai2018RapidAW}. In this area, a siamese network was proposed in~\cite{koch2015siamese} that consists of twin networks sharing weights, where each network is respectively fed with a support image and a query. The distance between the query and its support is naturally learned by a logistic regression. This matching strategy captures inherent variety between support and query regardless of their categories.  In the realm of matching framework, subsequent works~\cite{santoro2016meta, vinyals2016matching, cai2018memory, yang2018learning, kang2018few, wang2018low} had focused on enhancing feature embedding, where one direction is to build memory modules to capture global contexts among the supports. A number of works~\cite{li2019DN4, lifchitz2019dense} exploit local descriptors to reap additional knowledge from limited data. In~\cite{kim2019egnn, gidaris2019generating} the authors introduced Graph Neural Network (GNN) to model relationship between different categories. In~\cite{li2019finding} the given entire support set was traversed to identify task-relevant features and to make metric learning in high-dimensional space more effective. Other works, such as~\cite{Sachin2017, finn2017model}, dedicate to learning a general agent to guide parameter optimization.

Until now,  few-shot learning has not achieved groundbreaking progress, which has mostly focused on the classification task but rarely on other important computer vision tasks such as semantic segmentation~\cite{dong2018few, Michaelis2018OneShotSI, Hu2019AttentionbasedMG}, human motion prediction~\cite{Gui2018FewShotHM} and object detection~\cite{Chen2018LSTDAL}. 
In~\cite{dong2018pami} unlabeled data was used and multiple modules were optimized alternately on images without box. However, the method may be misled by incorrect detection in weak supervision and requires re-training for a new category. In LSTD~\cite{Chen2018LSTDAL} the authors proposed a novel few-shot object detection framework that can transfer knowledge from one large dataset to another smaller dataset, by minimizing the gap of classifying posterior probability between the source domain and the target domain. This method, however, strongly depends on the source domain and is hard to extend to very different scenarios. Recently, several other works for few-shot detection~\cite{Chen2018LSTDAL, kang2018few, karlinsky2019repmet, yan2019metarcnn} have been proposed but they learn category-specific embeddings and require to be fine-tuned for novel categories. 

Our work is motivated by the research line pioneered by the matching network~\cite{koch2015siamese}. We propose a general few-shot object detection network that learns the matching metric between image pairs based on the Faster R-CNN framework equipped with our novel attention RPN and multi-relation detector trained using our contrastive training strategy.

\section{FSOD: A Highly-Diverse Few-Shot Object Detection Dataset}
\label{section:dataset}

\begin{figure*}[!ht]
\centering
\includegraphics[width=13.5cm]{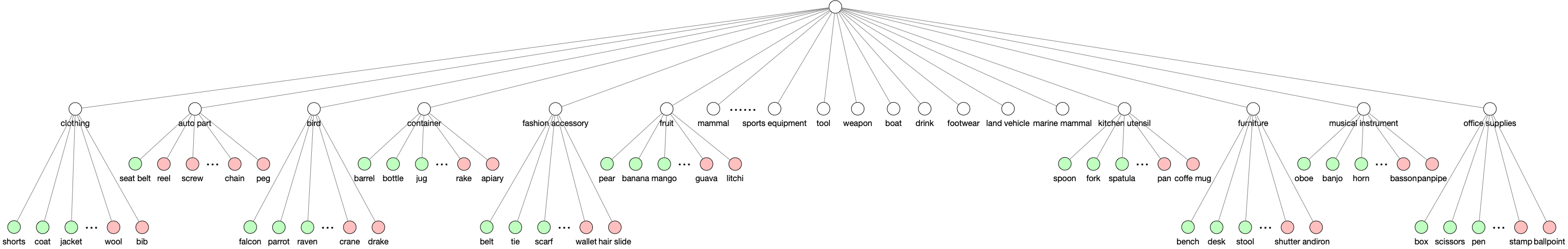}
\caption{Dataset label tree. The ImageNet categories (red circles) are merged with Open Image categories (green circles) where the superclasses are adopted.}
\label{fig:dataset_tree}
\vspace{-0.5cm}
\end{figure*}

The key to few-shot learning lies in the generalization ability of the pertinent model when presented with novel categories. Thus, a high-diversity dataset with a large number of object categories is necessary for training a general model that can detect unseen objects and for performing convincing evaluation as well. However, existing datasets~\cite{lin2014microsoft, Geiger2012CVPR,everingham2010pascal,OpenImages,krishna2017visual} contain very limited categories and they are not designed in the few-shot evaluation setting. Thus we build a new few-shot object detection dataset. 

\begin{figure}[!t]
\centering
\includegraphics[width=7cm]{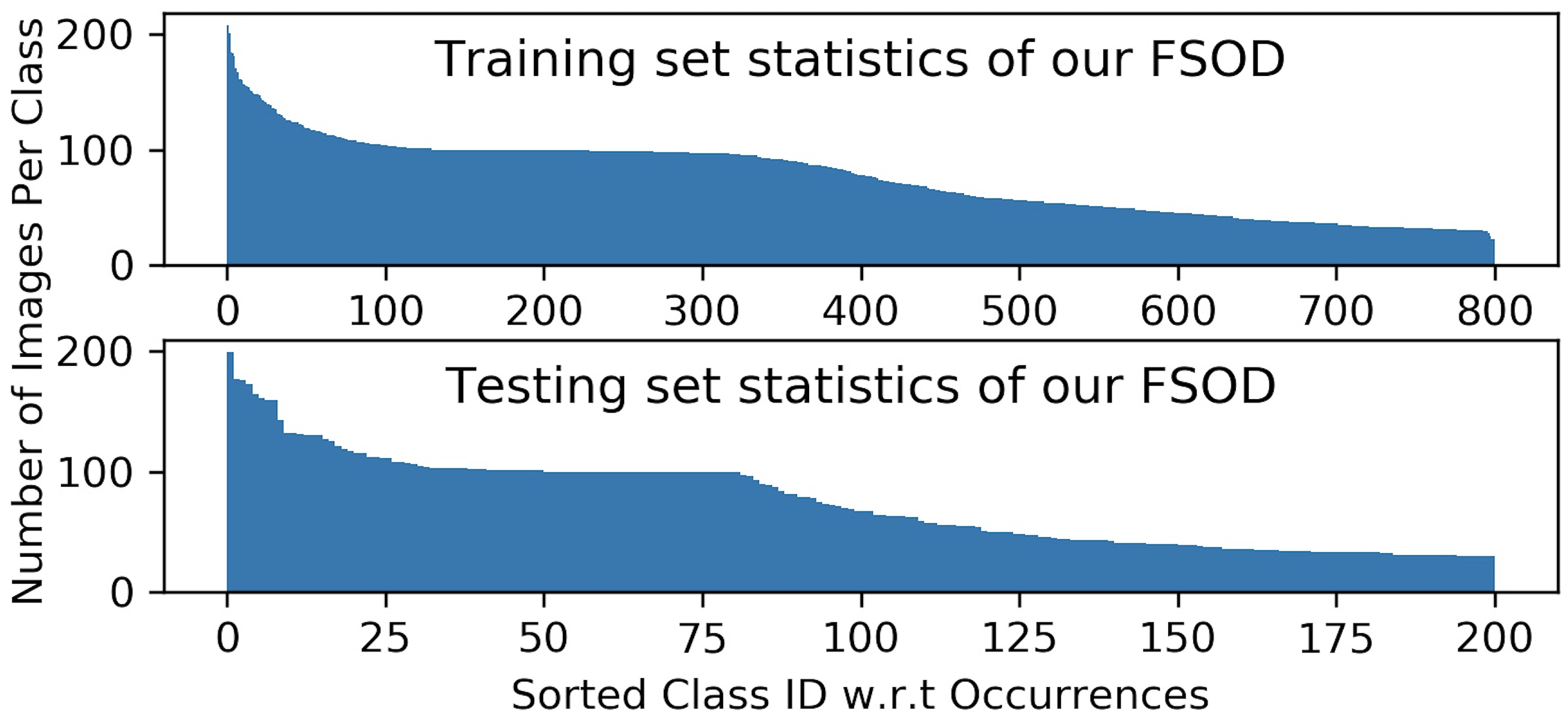} 
\caption{The dataset statistics of FSOD. The category image number are distributed almost averagely. Most classes (above 90\%) has small or moderate amount of images (in [22, 108]), and the most frequent class still has no more than 208 images.}
\label{fig:dataset_hist}
\vspace{-0.45cm}
\end{figure}

\noindent {\bf Dataset Construction.} We build our dataset from existing large-scale object detection datasets for supervised learning \ie~\cite{OpenImages, deng2009imagenet}. These datasets, however, cannot be used directly, due to 1) the label system of different datasets are inconsistent where some objects with the same semantics are annotated with different words in the datasets; 2) large portion of the existing annotations are noisy due to inaccurate and missing labels, duplicate boxes, objects being too large; 3) their train/test split contains the same categories, while for the few-shot setting we want the train/test sets to contain different categories in order to evaluate its generality on unseen categories.

To start building the dataset, we first summarize a label system from~\cite{OpenImages, deng2009imagenet}. We merge the leaf labels in their original label trees, by grouping those in the same semantics (e.g., ice bear and polar bear) into one category, and removing semantics that do not belong to any leaf categories. Then, we remove the images with bad label quality and those with boxes of improper size. Specifically, removed images have boxes smaller than 0.05\% of the image size which are usually in bad visual quality and unsuitable to serve as support examples. Next, we follow the few-shot learning setting to split our data into training set and test set without overlapping categories. We construct the training set with categories in MS COCO dataset~\cite{lin2014microsoft} in case researchers prefer a pretraining stage. We then split the test set which contains 200 categories by choosing those with the largest distance with existing training categories, where the distance is the shortest path that connects the meaning of two phrases in the is-a taxonomy~\cite{miller1995wordnet}. The remaining categories are merged into the training set that in total contains 800 categories. In all, we construct a dataset of 1000 categories with unambiguous category split for training and testing, where 531 categories are from ImageNet dataset~\cite{deng2009imagenet} and 469 from Open Image dataset~\cite{OpenImages}.

\noindent {\bf Dataset Analysis.} Our dataset is specifically designed for few-shot learning and for evaluating the generality of a model on novel categories, which contains 1000 categories with 800/200 split for training and test set respectively, around 66,000 images and 182,000 bounding boxes in total. Detailed statistics are shown in Table~\ref{dataset_summery_table} and Fig.~\ref{fig:dataset_hist}. Our dataset has the following properties: 

{\em High diversity in categories:} Our dataset contains 83 parent semantics, such as mammal, clothing, weapon, etc, which are further split to 1000 leaf categories. Our label tree is shown in Fig.~\ref{fig:dataset_tree}. Due to our strict dataset split, our train/test sets contain images of very different semantic categories thus presenting challenges to models to be evaluated.

{\em Challenging setting:} Our dataset contains objects with large variance on box size and aspect ratios, consisting of 26.5\% images with no less than three objects in the test set. Our test set contains a large number of boxes of categories {\em not} included in our label system, thus presenting great challenges for a few-shot model.

\begin{figure*}[!ht]
\centering
\includegraphics[width=0.83\linewidth]{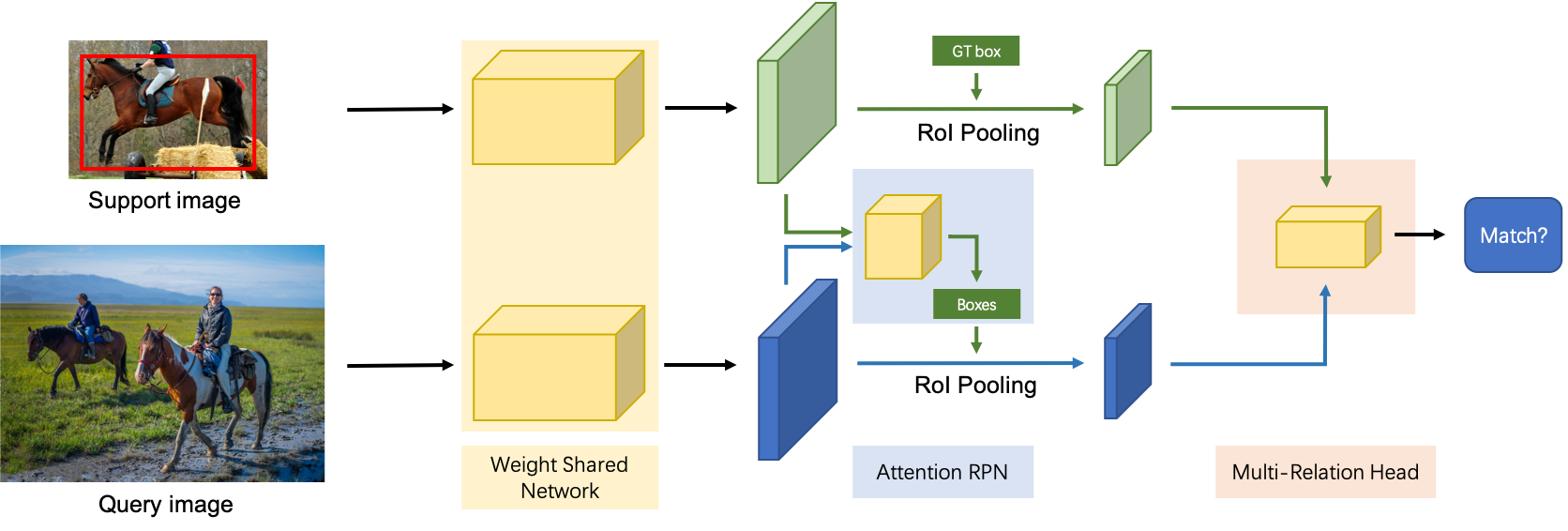}
\caption{Network architecture. The query image and support image are processed by the weight-shared network. The attention RPN module filters out object proposals in other categories by focusing on the given support category. The multi-relation detector then matches the query proposals and the support object. For the $N$-way training, we extend the network by adding $N-1$ support branches where each branch has its own attention RPN and multi-relation detector with the query image. For $K$-shot training, we obtain all the support feature through the weight-shared network and use the average feature across all the supports belonging to the same category as its support feature.}
\label{fig:framework}
\vspace{-0.3cm}
\end{figure*}

\begin{table}
\begin{center}
{\small
\begin{tabular}{|l|c|c|}
\hline
 & Train & Test \\
\hline\hline
No. Class & 800 & 200 \\
No. Image & 52350 & 14152 \\
No. Box & 147489 & 35102 \\
Avg No. Box / Img  & 2.82 & 2.48 \\ 
Min No. Img / Cls  & 22 & 30 \\ 
Max No. Img / Cls  & 208 & 199 \\ 
Avg No. Img / Cls  & 75.65 & 74.31 \\ 
Box Size & [6, 6828] & [13, 4605] \\
Box Area Ratio & [0.0009, 1] & [0.0009, 1] \\
Box W/H Ratio & [0.0216, 89] & [0.0199, 51.5] \\
\hline
\end{tabular}}
\end{center}
\vspace{-0.2cm}
\caption{Dataset Summary. Our dataset is diverse with large variance in box size and aspect ratio.}
\label{dataset_summery_table}
\vspace{-0.45cm}
\end{table}

Although our dataset has a large number of categories, the number of training images and boxes are much less than other large-scale benchmark datasets such as MS COCO dataset, which contains 123,287 images and around 886,000 bounding boxes. Our dataset is designed to be compact while effective for few-shot learning.

\section{Our Methodology}

In this section, we first define our task of few-shot detection, followed by a detailed description of our novel few-shot object detection network.

\subsection{Problem Definition}
Given a support image $s_{c}$ with a close-up of the target object and a query image $q_{c}$ which potentially contains objects of the support category $c$, the task is to find all the target objects belonging to the support category in the query and label them with tight bounding boxes. If the support set contains $N$ categories and $K$ examples for each category, the problem is dubbed $N$-way $K$-shot detection. 

\subsection{Deep Attentioned Few-Shot Detection}


We propose a novel attention network that learns a general matching relationship between the support set and queries on both the RPN module and the detector. Fig.~\ref{fig:framework} shows the overall architecture of our network.
Specifically, we build a weight-shared framework that consists of multiple branches, where one branch is for the query set and the others are for the support set (for simplicity, we only show one support branch in the figure). The query branch of the weight-shared framework is a Faster R-CNN network, which contains RPN and detector. We utilize this framework to train the matching relationship between support and query features, in order to make the network learn general knowledge among the same categories. Based on the framework, we introduce a novel attention RPN and detector with multi-relation modules to produce an accurate parsing between support and potential boxes in the query. 

\vspace{-0.1in}
\subsubsection{Attention-Based Region Proposal Network}

\begin{figure}[!t]
\centering
\includegraphics[width=0.81\linewidth]{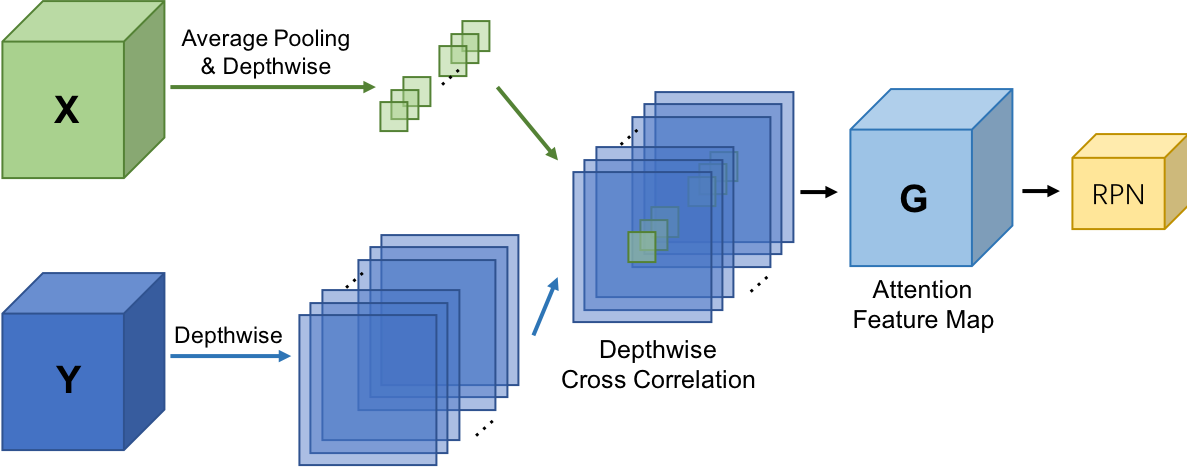}
\caption{Attention RPN. The support feature is average pooled to a $1 \times 1 \times C$ vector. Then the depth-wise cross correlation with the query feature is computed whose output is used as attention feature to be fed into RPN for generating proposals.}
\label{fig:rpn}
\vspace{-0.4cm}
\end{figure}

In few-shot object detection, RPN is useful in producing potentially relevant boxes for facilitating the following task of detection. Specifically, the RPN should not only distinguish between objects and non-objects but also filter out negative objects not belonging to the support category. However, without any support image information, the RPN will be aimlessly active in every potential object with high objectness score even though they do not belong to the support category, thus burdening the subsequent classification task of the detector with a large number of irrelevant objects. To address this problem, we propose the attention RPN (Fig.~\ref{fig:rpn}) which uses support information to enable filtering out most background boxes and those in non-matching categories. Thus a smaller and more precise set of candidate proposals is generated with high potential containing target objects.

We introduce support information to RPN through the attention mechanism to guide the RPN to produce relevant proposals while suppressing proposals in other categories. Specifically, we compute the similarity between the feature map of support and that of the query in a depth-wise manner. The similarity map then is utilized to build the proposal generation. In particular, we denote the support features as $X\in t^{S \times S \times C}$ and feature map of the query as $Y\in t^{H\times W \times C}$, the similarity is defined as
\vspace{-0.4cm}

\begin{equation*}
\begin{split}
    &\mathbf{G}_{h,w,c} = \sum_{i,j} X_{i,j,c} \cdot  Y_{h+i-1,w+j-1,c}, 
\quad i,j\in \{1,...,S\} 
\end{split}
\label{equ:attentionrpn}
\end{equation*}

\noindent where ${\mathbf G}$ is the resultant attention feature map. Here the support features $X$ is used as the kernel to slide on the query feature map~\cite{bertinetto2016fully, lu2018class} in a depth-wise cross correlation manner~\cite{li2018siamrpn++}. 
In our work, we adopt the features of top layers to the RPN model, \ie the res4\_6 in ResNet50. We find that a kernel size of $S=1$ performs well in our case. This fact is consistent with~\cite{ren2015faster} that global feature can provide a good object prior for objectness classification. In our case, the kernel is calculated by averaging on the support feature map. The attention map is processed by a $3\times 3$ convolution followed by the objectiveness classification layer and box regression layer. The attention RPN with loss $L_{rpn}$ is trained jointly with the network as in~\cite{ren2015faster}.

\begin{figure}[!t]
\centering
\includegraphics[width=5.8cm]{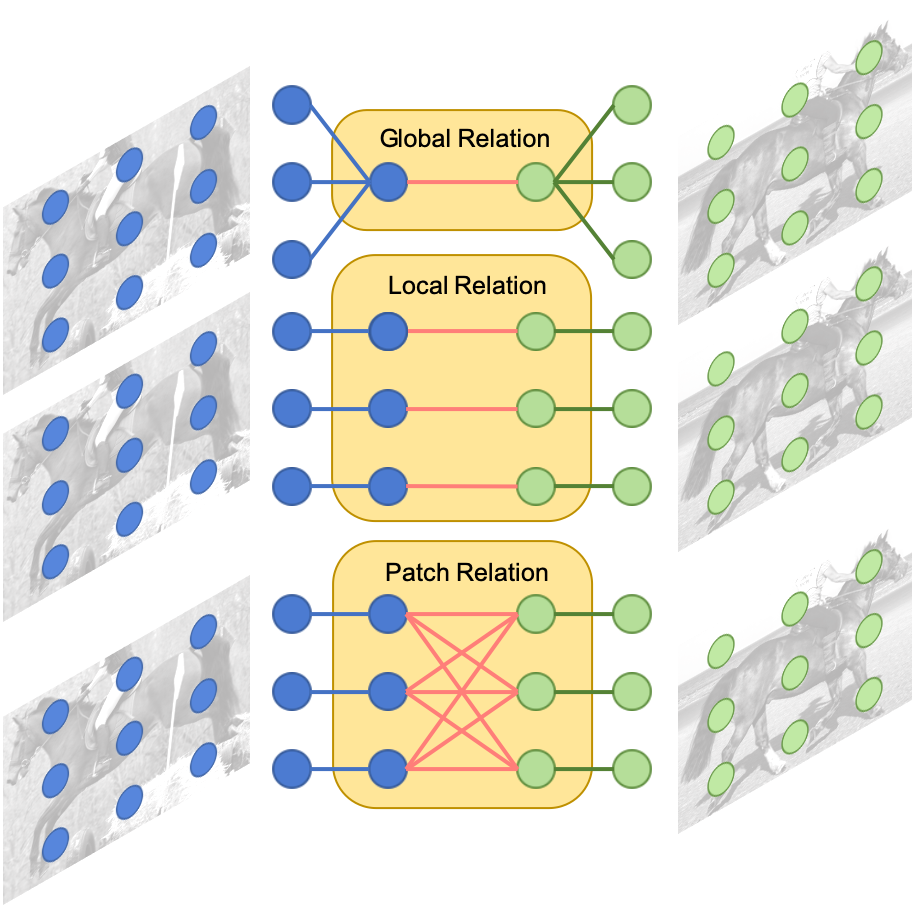}
\caption{Multi-Relation Detector. Different relation heads model different relationships between the query and support image. The global relation head uses global representation to match images; local relation head captures pixel-to-pixel matching relationship; patch relation head models one-to-many pixel relationship.}
\label{fig:head}
\vspace{-0.4cm}
\end{figure}

\subsubsection{Multi-Relation Detector}

In an R-CNN framework, an RPN module will be followed by a detector whose important role is re-scoring proposals and class recognition. Therefore, we want a detector to have a strong discriminative ability to distinguish different categories. To this end, we propose a novel multi-relation detector to effectively measure the similarity between proposal boxes from the query and the support objects, see Fig.~\ref{fig:head}. The detector includes three attention modules, which are respectively the {\bf global-relation head} to learn a deep embedding for global matching, the {\bf local-correlation head} to learn the pixel-wise and depth-wise cross correlation between support and query proposals and the {\bf patch-relation head} to learn a deep non-linear metric for patch matching. We experimentally show that the three matching modules can complement each other to produce higher performance. Refer to the supplemental material for implementation details of the three heads. 

\noindent {\bf Which relation heads do we need?}
We follow the $N$-way $K$-shot evaluation protocol proposed in RepMet~\cite{schwartz2018repmet} to evaluate our relation heads and other components. Table~\ref{table:head} shows the ablation study of our proposed multi-relation detector under the naive 1-way 1-shot training strategy and 5-way 5-shot evaluation on the FSOD dataset. We use the same evaluation setting hereafter for all ablation studies on the FSOD dataset. For individual heads, the local-relation head performs best on both $AP_{50}$ and $AP_{75}$ evaluations. Surprisingly, the patch-relation head performs worse than other relation heads, although it models more complicated relationship between images. We believe that the complicated relation head makes the model difficult to learn. When combining any two types of relation head, we obtain better performance than that of individual head. By combining all relation heads, we obtain the full multi-relation detector and achieve the best performance, showing that the three proposed relation heads are complementary to each other for better differentiation of targets from non-matching objects. All the following experiments thus adopt the full multi-relation detector.

\begin{table}
\begin{center}{\small
\begin{tabular}{|c|c|c|c|c|}
\hline
Global R        & Local R        & Patch R        & $AP_{50}$ & $AP_{75}$ \\ 
\hline\hline
\cmark &           &           & 47.7    & 34.0    \\ 
          & \cmark &           & 50.5    & 35.9    \\ 
          &           & \cmark & 45.1    & 32.8    \\ 
\cmark &           & \cmark & 49.6    & 35.9    \\ 
          & \cmark & \cmark & 53.8    & 38.0    \\ 
\cmark & \cmark &           & 54.6    & 38.9    \\ 
\cmark & \cmark & \cmark & {\bf 55.0}    & {\bf 39.1}    \\ 
\hline
\end{tabular}}
\end{center}
\vspace{-0.1in}
\caption{Experimental results for different relation head combinations in the 1-way 1-shot training strategy.}
\label{table:head}
\vspace{-0.2in}
\end{table}

\subsection{Two-way Contrastive Training Strategy}

A naive training strategy is matching the same category objects by constructing a training pair ($q_{c}$, $s_{c}$) where the query image $q_{c}$ and support image $s_{c}$ are both in the same $c$-th category object. However a good model should not only match the same category objects but also distinguish different categories. 
For this reason, we propose a novel 2-way contrastive training strategy.

\begin{figure}[!t]
\centering
\includegraphics[width=0.9\linewidth]{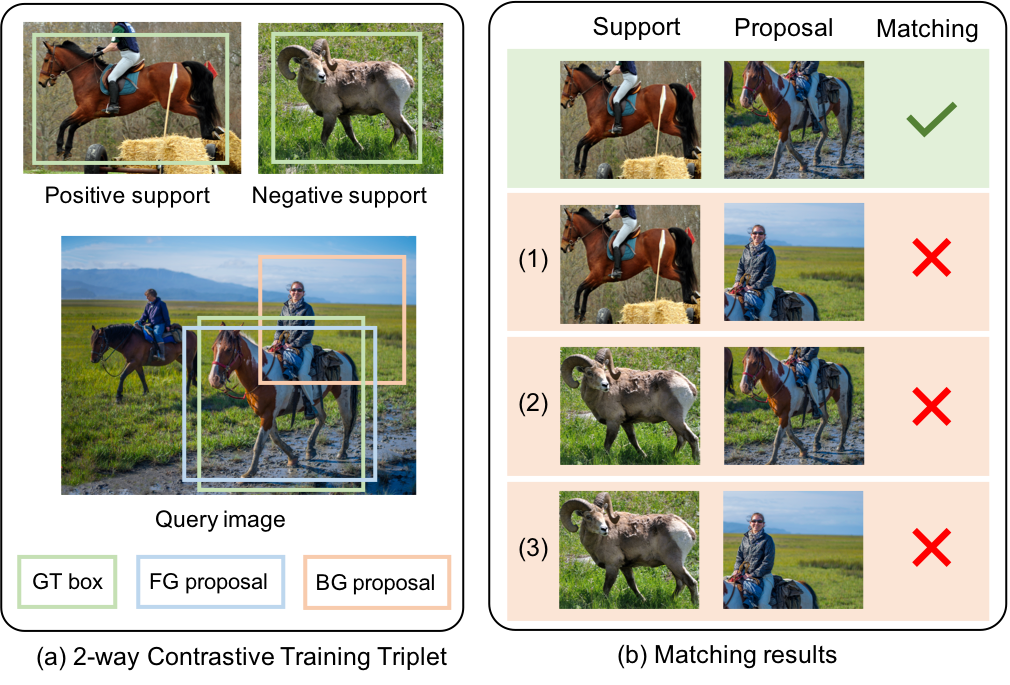}
\caption{The 2-way contrastive training triplet and different matching results. Only the positive support has the same category with the target ground truth in the query image. The matching pair consists of the positive support and foreground proposal, and the non-matching pair has three categories: (1) positive support and background proposal, (2) negative support and foreground proposal and (3) negative support and background proposal.}
\label{fig:training}
\vspace{-0.45cm}
\end{figure}

According to the different matching results in Fig.~\ref{fig:training}, we propose the 2-way contrastive training to match the same category while distinguishing different categories. We randomly choose one query image $q_{c}$, one support image $s_{c}$ containing the same $c$-th category object and one other support image $s_{n}$ containing a different $n$-th category object, to construct the training triplet ($q_{c}$, $s_{c}$, $s_{n}$), where $c \neq n$. In the training triplet, only the $c$-th category objects in the query image are labeled as foreground while all other objects are treated as background. 

During training, the model learns to match every proposal generated by the attention RPN in the query image with the object in the support image. Thus the model learns to not only match the same category objects between ($q_{c}$, $s_{c}$) but also distinguish objects in different categories between ($q_{c}$, $s_{n}$). However, there are a massive amount of background proposals which usually dominate the training, especially with negative support images. For this reason, we balance the ratio of these matching pairs between query proposals and supports. We keep the ratio as 1:2:1 for the foreground proposal and positive support pairs $(p_f, s_p)$, background proposal and positive support pairs $(p_b, s_p)$, and proposal (foreground or background) and negative support pairs $(p, s_n)$. We pick all $N$ $(p_f, s_p)$ pairs and select top $2N$ $(p_b, s_p)$ pairs and top $N$ $(p, s_n)$ pairs respectively according to their matching scores and calculate the matching loss on the selected pairs. During training, we use the multi-task loss on each sampled proposal as $L = L_{matching} + L_{box}$ with the bounding-box loss $L_{box}$ as defined in~\cite{girshick2015fast} and the matching loss being the binary cross-entropy.


\noindent {\bf Which training strategy is better?} Refer to Table~\ref{table:training}.
We train our model with the 2-way 1-shot contrastive training strategy and obtain 7.9\% $AP_{50}$ improvement compared with the naive 1-way 1-shot training strategy, which indicates the importance in learning how to distinguish different categories during training. With 5-shot training, we achieve further improvement which was also verified in~\cite{snell2017prototypical} that few-shot training is beneficial to few-shot testing. It is straightforward to extend our 2-way training strategy to multi-way training strategy. However, from Table~\ref{table:training}, the 5-way training strategy does not produce better performance than the 2-way training strategy. We believe that only one negative support category suffices in training the model for distinguishing different categories. Our full model thus adopts the 2-way 5-shot contrastive training strategy.

\noindent {\bf Which RPN is better?}
We evaluate our attention RPN on different evaluation metrics. To evaluate the proposal quality, we first evaluate the recall on top 100 proposals over 0.5 IoU threshold of the regular RPN and our proposed attention RPN. Our attention RPN exhibits better recall performance than the regular RPN (0.9130 \vs 0.8804). We then evaluate the average best overlap ratio (ABO~\cite{uijlings2013selective}) across ground truth boxes for these two RPNs. The ABO of attention RPN is 0.7282 while the same metric of regular RPN is 0.7127. These results indicate that the attention RPN can generate more high-quality proposals. 

Table~\ref{table:training} further compares models with attention RPN and those with the regular RPN in different training strategies. The model with attention RPN consistently performs better than the regular RPN on both $AP_{50}$ and $AP_{75}$ evaluation. The attention RPN produces 0.9\%/2.0\% gain in the 1-way 1-shot training strategy and 2.0\%/2.1\% gain in the 2-way 5-shot training strategy on the $AP_{50}$/$AP_{75}$ evaluation. These results confirm that our attention RPN generates better proposals and benefits the final detection prediction. The attention RPN is thus adopted in our full model. 

\section{Experiments}

In the experiments, we compare our approach with state-of-the-art (SOTA) methods on different datasets. We typically train our full model on FSOD training set 
and directly evaluate on these datasets. For fair comparison with other methods, we may discard training on FSOD and adopt the same train/test setting as these methods. In these cases, we use a multi-way\footnote{The fine-tuning stage benefits from more ways during the multi-way training, so we use as many ways as possible to fill up the GPU memory.} few-shot training in the fine-tuning stage with more details to be described. 

\begin{table}
\begin{center}{\small
\begin{tabular}{|c|c|c|c|c|}
\hline
Training Strategy & Attention RPN  & $AP_{50}$ & $AP_{75}$ \\
\hline\hline
1-way 1-shot      &                & 55.0    & 39.1    \\
1-way 1-shot      &\cmark          & 55.9    & 41.1    \\
2-way 1-shot      &                & 63.8    & 42.9    \\
2-way 5-shot      &                & 65.4    & 43.7    \\
2-way 5-shot      &\cmark          & {\bf 67.5}    & {\bf 46.2}    \\
5-way 5-shot      &\cmark          & 66.9    & 45.6    \\
\hline
\end{tabular}}
\end{center}
\vspace{-0.1in}
\caption{Experimental results for training strategy and attention RPN with the multi-relation detector.} 
\label{table:training}
\vspace{-0.15in}
\end{table}

\subsection{Training Details}

Our model is trained end-to-end on 4 Tesla P40 GPUs using SGD with a batch size of 4 (for query images).
The learning rate is 0.002 for the first 56000 iterations and 0.0002 for later 4000 iterations. We observe that pre-training on ImageNet~\cite{deng2009imagenet} and MS COCO~\cite{lin2014microsoft} can provide stable low-level features and lead to a better converge point. Given this, we by default train our model from the pre-trained ResNet50 on~\cite{lin2014microsoft, deng2009imagenet} unless otherwise stated.   
During training, we find that more training iterations may damage performance, where too many training iterations make the model over-fit to the training set. We fix the weights of Res1-3 blocks and only train high-level layers to utilize low-level basic features and avoid over-fitting. The shorter side of the query image is resized to 600 pixels; the longer side is capped at 1000. The support image is cropped around the target object with 16-pixel image context, zero-padded and then resized to a square image of $320 \times 320$. For few-shot training and testing, we fuse feature by averaging the object features with the same category and then feed them to the attention RPN and the multi-relation detector. We adopt the typical metrics~\cite{lin2017focal}, i.e. \emph{AP}, \emph{AP$_{50}$} and \emph{AP$_{75}$} for evaluation.

\subsection{Comparison with State-of-the-Art Methods}

\subsubsection{ImageNet Detection dataset}
In Table~\ref{table:imagenet}, we compare our results with those of LSTD~\cite{Chen2018LSTDAL} and RepMet~\cite{schwartz2018repmet} on the challenging ImageNet based 50-way 5-shot detection scenario. For fair comparison, we use their evaluation protocol and testing dataset and we use the same MS COCO training set to train our model. We also use soft-NMS~\cite{Bodla2017Soft} as RepMet during evaluation. Our approach produces 1.7\% performance gain compared to the state-of-the-art (SOTA) on the $AP_{50}$ evaluation. 

To show the generalization ability of our approach, we directly apply our model trained on FSOD dataset 
on the test set and we obtain 41.7\% on the $AP_{50}$ evaluation which is surprisingly better than our fine-tuned model (Table~\ref{table:imagenet}). It should be noted that our model trained on FSOD dataset can be directly applied on the test set without fine-tuning to achieve SOTA performance. Furthermore, although our model trained on FSOD dataset has a slightly better $AP_{50}$ performance than our fine-tuned model on the MS COCO dataset, our model surpasses the fine-tuned model by 6.4\% on the $AP_{75}$ evaluation, which shows that our proposed FSOD dataset significantly benefits few-shot object detection. With further fine-tuning our FSOD trained model on the test set,  our model achieves the best performance, while noting that our method without fine-tuning already works best compared with SOTA.

\begin{table}
\begin{center}
{\small
\begin{tabular}{|c|c|c|c|c|}
\hline
Method & dataset & fine-tune  & $AP_{50}$ & $AP_{75}$ \\
\hline\hline
LSTD~\cite{Chen2018LSTDAL}   & COCO    & \cmark$^{ImageNet}$  & 37.4      &     -      \\
RepMet~\cite{karlinsky2019repmet} & COCO    & \cmark$^{ImageNet}$ & 39.6      &     -      \\
Ours   & COCO    &    \cmark$^{ImageNet}$      & 41.3      & 21.9      \\
Ours   & FSOD$^\dag$    &    \xmark      & 41.7      & 28.3      \\
Ours   & FSOD$^\dag$    &    \cmark$^{ImageNet}$      & {\bf 44.1}      & {\bf 31.0}      \\
\hline
\end{tabular}
}
\end{center}
\vspace{-0.1in}
\caption{Experimental results on ImageNet Detection dataset for 50 novel categories with 5 supports. $^\dagger$ means that the testing categories are removed from FSOD training dataset. \cmark$^{ImageNet}$ means the model is fine-tuned on ImageNet Detection dataset.}
\label{table:imagenet}
\vspace{-0.1in}
\end{table}

\begin{table}
\begin{center}
{\small
\begin{tabular}{|c|c|c|c|c|c|}
\hline
Method & dataset & fine-tune  & $AP$ & $AP_{50}$ & $AP_{75}$ \\
\hline\hline
FR~\cite{kang2018few} & COCO    & \cmark$^{coco}$ & 5.6 & 12.3      & 4.6       \\
Meta~\cite{yan2019metarcnn}& COCO    & \cmark$^{coco}$ & 8.7 &   19.1    &  6.6     \\
Ours    & COCO    & \cmark$^{coco}$ & 11.1 &       20.4    &      10.6     \\
Ours    & FSOD$^\dag$    &   \xmark         & {\bf 16.6} &    {\bf 31.3}   &   {\bf 16.1}  \\
\hline
\end{tabular}
}
\end{center}
\vspace{-0.1in}
\caption{Experimental results on MS COCO minival set for 20 novel categories with 10 supports.
$^\dagger$ means that the testing categories are removed from FSOD training dataset. \cmark$^{coco}$ means the model is fine-tuned on MS COCO dataset.}
\label{table:coco}
\vspace{-0.15in}
\end{table}

\vspace{-0.1in}

\subsubsection{MS COCO dataset}
\label{coco}

In Table~\ref{table:coco}, we compare our approach\footnote{Since Feature Reweighting and Meta R-CNN are evaluated on MS COCO, in this subsection we discard pre-training on~\cite{lin2014microsoft} for fair comparison to follow the same experimental setting as described.} 
with Feature Reweighting~\cite{kang2018few} and Meta R-CNN~\cite{yan2019metarcnn} on MS COCO minival set. 
We follow their data split and use the same evaluation protocol: we set the 20 categories included in PASCAL VOC as novel categories for evaluation, and use the rest 60 categories in MS COCO as training categories. Our fine-tuned model with the same MS COCO training dataset outperforms Meta R-CNN by 2.4\%/1.3\%/4.0\% on $AP$/$AP_{50}$/$AP_{75}$ metrics. This demonstrates the strong learning and generalization ability of our model, as well as that, in the few-shot scenario, learning general matching relationship is more promising than the attempt to learn category-specific embeddings~\cite{kang2018few, yan2019metarcnn}.
Our model trained on FSOD achieves more significant improvement of 7.9\%/12.2\%/9.5\% on $AP$/$AP_{50}$/$AP_{75}$ metrics. Note that our model trained on FSOD dataset are directly applied on the novel categories without any further fine-tuning while all other methods use 10 supports for fine-tuning to adapt to the novel categories. Again, without fine-tuning our FSOD-trained model already works the best among SOTAs. 

\begin{figure*}[!ht]
\centering
\includegraphics[width=0.82\linewidth]{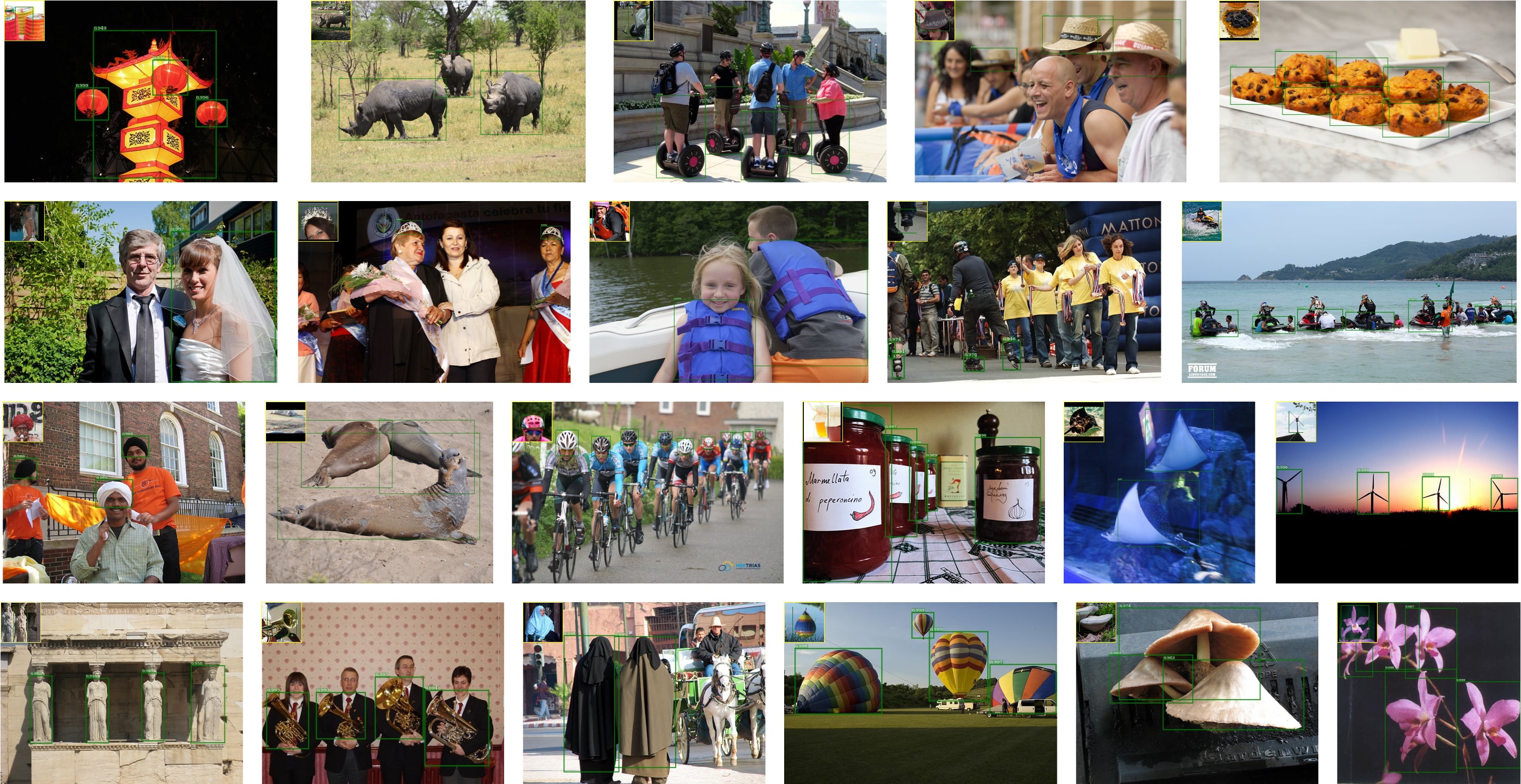}
\caption{Qualitative 1-shot detection results of our approach on FSOD test set. Zoom in the figures for more visual details.}
\vspace{-0.1cm}
\label{fig:vis}
\vspace{-0.1cm}
\end{figure*}

\begin{figure*}[!ht]
\centering
\includegraphics[width=0.83\linewidth]{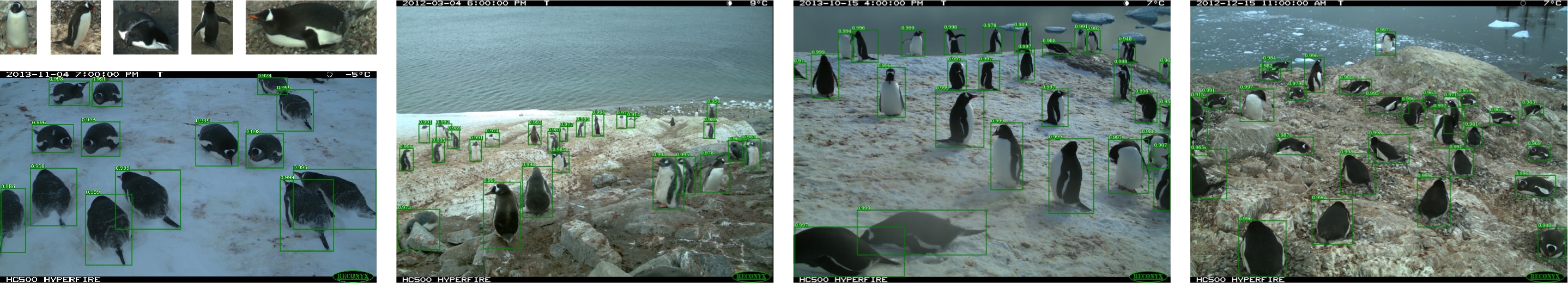}
\caption{Our application results on the penguin dataset~\cite{Arteta16}. Given 5 penguin images as support, our approach can detect all penguins in the wild in the given query image.}
\label{fig:penguin}
\vspace{-0.2in}
\end{figure*}

\subsection{Realistic Applications}
\label{applications}

We apply our approach in different real-world application scenarios to demonstrate its generalization capability. Fig.~\ref{fig:vis} shows qualitative 1-shot object detection results on novel categories in our test set. We further apply our approach on the wild penguin detection~\cite{Arteta16} and show sample qualitative 5-shot object detection results in Fig.~\ref{fig:penguin}.

\begin{table}
\begin{center}{\small
\begin{tabular}{|c|c|c|c|c|c|c|c|}
\hline
Method & FSOD pretrain & fine-tune  & $AP_{50}$ & $AP_{75}$ \\
\hline\hline
FRCNN~\cite{ren2015faster} & \xmark & \cmark$^{fsod}$ & 11.8  & 6.7 \\ 
FRCNN~\cite{ren2015faster} & \cmark & \cmark$^{fsod}$ & 23.0  & 12.9 \\
LSTD~\cite{Chen2018LSTDAL}  & \cmark & \cmark$^{fsod}$ & 24.2 & 13.5   \\
Ours                     & trained directly  & \xmark          & {\bf 27.5}  & {\bf 19.4} \\
\hline
\end{tabular}}
\end{center}
\vspace{-0.1in}
\caption{Experimental results on FSOD test set for 200 novel categories with 5 supports evaluated in novel category detection. \cmark$^{fsod}$ means the model is fine-tuned on FSOD dataset.} 
\label{table:fsod}
\vspace{-0.2in}
\end{table}

\noindent {\bf Novel Category Detection.}
Consider this common real-world application scenario: given a massive number of images in a photo album or TV drama series without any labels, the task is to annotate a novel target object (e.g., a rocket) in the given massive collection without knowing which images contain the target object, which can be in different sizes and locations if present.
In order to reduce manual labor, one solution is to manually find a small number of images containing the target object, annotate them, and then apply our method to automatically annotate the rest in the image collection. Following this setting, we perform the evaluation as follows: We mix all test images of FSOD dataset, and for each object category, we pick 5 images that contain the target object to perform this novel category object detection in the entire test set. 
	Note that different from the standard object detection evaluation, in this evaluation, the model evaluates every category separately and has no knowledge of the complete categories. 

We compare with LSTD~\cite{Chen2018LSTDAL} which needs to be trained on novel categories by transferring knowledge from the source to target domain. Our method, however, can be applied to detect object in novel categories \textbf{without any further re-training or fine-tuning}, which is fundamentally different from LSTD. To compare empirically, we adjust LSTD to base on Faster R-CNN and re-train it on 5 fixed supports for each test category separately in a fair configuration. Results are shown in Table~\ref{table:fsod}. Our method outperforms LSTD by 3.3\%/5.9\% and its backbone Faster R-CNN by 4.5\%/6.5\% on all 200 testing categories on $AP_{50}/AP_{75}$ metrics. More specifically, without pre-training on our dataset, the performance of Faster R-CNN significantly drops. 
Note that because the model only knows the support category, the fine-tuning based models need to train every category separately which is time-consuming.


\renewcommand{\thefootnote}{2}

\noindent {\bf Wild Car Detection.}
We apply our method\footnote{We also discard the MS COCO pretraining in this experiment.} to wild car detection on KITTI~\cite{Geiger2012CVPR} and Cityscapes~\cite{Cordts2016Cityscapes} datasets which are urban scene datasets for driving applications, where the images are captured by car-mounted video cameras. We evaluate the performance of \textit{Car} category on KITTI training set with 7481 images and Cityscapes validation set with 500 images. 
DA Faster R-CNN~\cite{chen2018domain} uses massively annotated data from source domains (KITTI/Cityscapes) and unlabeled data from target domains (Cityscapes/KITTI) to train the domain adaptive Faster R-CNN, and evaluated the performance on target domains. Without any further re-training or fine-tuning, our model with 10-shot supports obtains comparable or even better $AP_{50}$ performance (37.0\% \vs 38.5\% on Cityscapes and 67.4\% \vs 64.1\% on KITTI) on the wild car detection task. Note that DA Faster R-CNN are specifically designed for the wild car detection task and they use much more training data in similar domains. 

\subsection{More Categories {\bf \vs} More Samples?}
\label{fsod_result}

Our proposed dataset has a large number of object categories but with few image samples in each category, which we claim is beneficial to few-shot object detection. To confirm this benefit, we train our model on MS COCO dataset, which has more than 115,000 images with only 80 categories. Then we train our model on FSOD dataset with different category numbers while keeping similar number of training image. Table~\ref{table:category} summarizes the experimental results, where we find that although MS COCO has the most training images but its model performance turns out to be the worst, while models trained on FSOD dataset have better performance as the number of categories incrementally increases while keeping similar number of training images, indicating that a limited number of categories with too many images can actually impede few-shot object detection,  while large number of categories can consistently benefit the task. Thus, we conclude that category diversity is essential to few-shot object detection.

\begin{table}
\begin{center}
{\small
\begin{tabular}{|c|c|c|c|c|}
\hline
Dataset & No. Class & No. Image       & $AP_{50}$ & $AP_{75}$ \\
\hline\hline
COCO~\cite{lin2014microsoft} & 80       &   115k        & 49.1    & 28.9    \\ 
FSOD & 300      &   26k        & 60.3    & 39.1    \\
FSOD & 500      &   26k        & 62.7    & 41.9    \\
FSOD & 800      &   27k        & {\bf 64.7}    & {\bf 42.6}    \\
\hline
\end{tabular}
}
\end{center}
\vspace{-0.1in}
\caption{Experimental results of our model on FSOD test set with different numbers of training categories and images in the 5-way 5-shot evaluation.}
\label{table:category}
\vspace{-0.2in}
\end{table}

\vspace{-0.2cm}

\section{Conclusion}
\vspace{-0.1cm}
We  introduce a novel few-shot object detection network with Attention-RPN, Multi-Relation Detectors and Contrastive Training strategy. We contribute a new FSOD which contains 1000 categories of various objects with high-quality annotations. Our model trained on FSOD can detect objects of novel categories requiring no pre-training or further network adaptation. Our model has been validated by extensive quantitative and qualitative results on different datasets. This paper contributes to few-shot object detection and we believe worthwhile and related future work can be spawn from our large-scale FSOD dataset and detection network with the above technical contributions.

\section*{Appendix A: Implementation Details of Multi-Relation Detector}

Given the support feature $f_s$ and query proposal feature $f_q$ with the size of $ 7 \times 7 \times C$, our multi-relation detector is implemented as follows. We use the sum of all matching scores from the three heads as the final matching scores. 

\vspace{-0.14in}

\paragraph{Global-Relation Head}
We concatenate $f_s$ and $f_q$ to the concatenated feature $f_c$ with the size of $ 7 \times 7 \times 2C$. Then we average pool $f_c$ to a $1\times 1\times 2C$ vector. We then use an MLP with two fully connected (fc) layers with ReLU and a final fc layer to process $f_c$ and generate matching scores. 

\vspace{-0.14in}

\paragraph{Local-Relation Head}
We first use a weight-shared $1\times 1\times C$ convolution to process $f_s$ and $f_q$ separately. Then we calculate the depth-wise similarity using the equation in Section 4.2.1 of the main paper with $S=H=W=7$. Then we use a fc layer to generate matching scores. 

 \vspace{-0.14in}
 
\paragraph{Patch-Relation Head}
We first concatenate $f_s$ and $f_q$ to the concatenated feature $f_c$ with the size of $ 7 \times 7 \times 2C$. Then $f_c$ is fed into the patch-relation module, whose structure is shown in Table~\ref{table:patch_relation}. All the convolution layers followed by ReLU and pooling layers in this module have zero padding to reduce the feature map size from $7\times7$ to $1\times1$. Then we use a fc layer to generate matching scores and a separate fc layer to generate bounding box predictions.

\section*{Appendix B: More Implementation Details}

\subsection*{B.1. Training and Fine-tuning details}

Here we show more details for the experiments in Section 5.2 of  the main paper.  

In Section 5.2, we follow other methods to train our model on MS COCO dataset~\cite{lin2014microsoft} and fine-tune on the target datasets. When we train our model on MS COCO, we remove the images with boxes smaller than the size of $32 \times 32$. Those boxes are usually in bad visual quality and hurt the training when they serve as support examples. When we fine-tune our model on the target datasets, we follow the same setting of other methods\cite{Chen2018LSTDAL, kang2018few, karlinsky2019repmet, yan2019metarcnn} for fair comparison. 
Specifically, LSTD~\cite{Chen2018LSTDAL} and RepMet~\cite{karlinsky2019repmet} use 5 support images per category where each image contains one or more object instances, and the Feature Reweighting~\cite{kang2018few} and Meta R-CNN~\cite{yan2019metarcnn} use a strict rule to adopt 10 object instances per category for fine-tuning.

\subsection*{B.2. Evaluation details}

There are two evaluation settings in the main paper. {\bf Evaluation setting 1:} The ablation experiments adopt the episode-based evaluation protocol defined in RepMet~\cite{karlinsky2019repmet}, where the setting is borrowed from the few-shot classification task~\cite{Sachin2017,vinyals2016matching}. There are 600 random evaluation episodes in total, which guarantee every image in the test set can be evaluated in a high probability. In each episode, for $N$-way $K$-shot evaluation, there are $K$ support images for each of the $N$ categories, and there are 10 query images for each category where each query image containing at least one instance belonging to this category. So there are $K \times N$ supports and $10 \times N$ query images in each episode. Note that all these categories and images are randomly chosen in each episode. {\bf Evaluation setting 2:} Other comparison experiments with baselines adopt the standard object detection evaluation protocol, which is a full-way, N-shot evaluation. 
	During evaluation, the support branches in our model can be discarded once the support features are attained, then the support features serve as model weights for the forward process. 




\begin{table}[!t]
\begin{center}
\begin{tabular}{|c|c|c|}
\hline
Type & Filter Shape & Stride/Padding \\
\hline\hline
 Avg Pool & 3x3x4096 & s1/p0 \\
 Conv & 1x1x512 & s1/p0 \\
 Conv & 3x3x512 & s1/p0 \\
 Conv & 1x1x2048 & s1/p0 \\
 Avg Pool & 3x3x2048 & s1/p0 \\
\hline
\end{tabular}
\end{center}
\vspace{-0.05in}
\caption{Architecture of the patch-relation module.}
\label{table:patch_relation}
\vspace{-0.15in}
\end{table}

\section*{Appendix D: FSOD Dataset Class Split}

Here we describe the training/testing class split in our proposed FSOD Dataset. This split was used in our experiments.

\subsection*{Training Class Split}

\noindent lipstick, sandal, crocodile, football helmet, umbrella, houseplant, antelope, woodpecker, palm tree, box, swan, miniskirt, monkey, cookie, scissors, snowboard, hedgehog, penguin, barrel, wall clock, strawberry, window blind, butterfly, television, cake, punching bag, picture frame, face powder, jaguar, tomato, isopod, balloon, vase, shirt, waffle, carrot, candle, flute, bagel, orange, wheelchair, golf ball, unicycle, surfboard, cattle, parachute, candy, turkey, pillow, jacket, dumbbell, dagger, wine glass, guitar, shrimp, worm, hamburger, cucumber, radish, alpaca, bicycle wheel, shelf, pancake, helicopter, perfume, sword, ipod, goose, pretzel, coin, broccoli, mule, cabbage, sheep, apple, flag, horse, duck, salad, lemon, handgun, backpack, printer, mug, snowmobile, boot, bowl, book, tin can, football, human leg, countertop, elephant, ladybug, curtain, wine, van, envelope, pen, doll, bus, flying disc, microwave oven, stethoscope, burrito, mushroom, teddy bear, nail, bottle, raccoon, rifle, peach, laptop, centipede, tiger, watch, cat, ladder, sparrow, coffee table, plastic bag, brown bear, frog, jeans, harp, accordion, pig, porcupine, dolphin, owl, flowerpot, motorcycle, calculator, tap, kangaroo, lavender, tennis ball, jellyfish, bust, dice, wok, roller skates, mango, bread, computer monitor, sombrero, desk, cheetah, ice cream, tart, doughnut, grapefruit, paddle, pear, kite, eagle, towel, coffee, deer, whale, cello, lion, taxi, shark, human arm, trumpet, french fries, syringe, lobster, rose, human hand, lamp, bat, ostrich, trombone, swim cap, human beard, hot dog, chicken, leopard, alarm clock, drum, taco, digital clock, starfish, train, belt, refrigerator, dog bed, bell pepper, loveseat, infant bed, training bench, milk, mixing bowl, knife, cutting board, ring binder, studio couch, filing cabinet, bee, caterpillar, sofa bed, violin, traffic light, airplane, closet, canary, toilet paper, canoe, spoon, fox, tennis racket, red panda, cannon, stool, zucchini, rugby ball, polar bear, bench, pizza, fork, barge, bow and arrow, kettle, goldfish, mirror, snail, poster, drill, tie, gondola, scale, falcon, bull, remote control, horn, hamster, volleyball, stationary bicycle, dishwasher, limousine, shorts, toothbrush, bookcase, baseball glove, computer mouse, otter, computer keyboard, shower, teapot, human foot, parking meter, ski, beaker, castle, mobile phone, suitcase, sock, cupboard, crab, common fig, missile, swimwear, saucer, popcorn, coat, plate, stairs, pineapple, parrot, fountain, binoculars, tent, pencil case, mouse, sewing machine, magpie, handbag, saxophone, panda, flashlight, baseball bat, golf cart, banana, billiard table, tower, washing machine, lizard, brassiere, ant, crown, oven, sea lion, pitcher, chest of drawers, crutch, hippopotamus, artichoke, seat belt, microphone, lynx, camel, rabbit, rocket, toilet, spider, camera, pomegranate, bathtub, jug, goat, cowboy hat, wrench, stretcher, balance beam, necklace, scoreboard, horizontal bar, stop sign, sushi, gas stove, tank, armadillo, snake, tripod, cocktail, zebra, toaster, frying pan, pasta, truck, blue jay, sink, lighthouse, skateboard, cricket ball, dragonfly, snowplow, screwdriver, organ, giraffe, submarine, scorpion, honeycomb, cream, cart, koala, guacamole, raven, drawer, diaper, fire hydrant, potato, porch, banjo, hammer, paper towel, wardrobe, soap dispenser, asparagus, skunk, chainsaw, spatula, ambulance, submarine sandwich, axe, ruler, measuring cup, scarf, squirrel, tea, whisk, food processor, tick, stapler, oboe, hartebeest, modem, shower cap, mask, handkerchief, falafel, clipper, croquette, house finch, butterfly fish, lesser scaup, barbell, hair slide, arabian camel, pill bottle, springbok, camper, basketball player, bumper car, wisent, hip, wicket, medicine ball, sweet orange, snowshoe, column, king charles spaniel, crane, scoter, slide rule, steel drum, sports car, go kart, gearing, tostada, french loaf, granny smith, sorrel, ibex, rain barrel, quail, rhodesian ridgeback, mongoose, red backed sandpiper, penlight, samoyed, pay phone, barber chair, wool, ballplayer, malamute, reel, mountain goat, tusker, longwool, shopping cart, marble, shuttlecock, red breasted merganser, shutter, stamp, letter opener, canopic jar, warthog, oil filter, petri dish, bubble, african crocodile, bikini, brambling, siamang, bison, snorkel, loafer, kite balloon, wallet, laundry cart, sausage dog, king penguin, diver, rake, drake, bald eagle, retriever, slot, switchblade, orangutan, chacma, guenon, car wheel, dandie dinmont, guanaco, corn, hen, african hunting dog, pajama, hay, dingo, meat loaf, kid, whistle, tank car, dungeness crab, pop bottle, oar, yellow lady's slipper, mountain sheep, zebu, crossword puzzle, daisy, kimono, basenji, solar dish, bell, gazelle, agaric, meatball, patas, swing, dutch oven, military uniform, vestment, cavy, mustang, standard poodle, chesapeake bay retriever, coffee mug, gorilla, bearskin, safety pin, sulphur crested cockatoo, flamingo, eider, picket fence, dhole, spaghetti squash, african elephant, coral fungus, pelican, anchovy pear, oystercatcher, gyromitra, african grey, knee pad, hatchet, elk, squash racket, mallet, greyhound, ram, racer, morel, drumstick, bovine, bullet train, bernese mountain dog, motor scooter, vervet, quince, blenheim spaniel, snipe, marmoset, dodo, cowboy boot, buckeye, prairie chicken, siberian husky, ballpoint, mountain tent, jockey, border collie, ice skate, button, stuffed tomato, lovebird, jinrikisha, pony, killer whale, indian elephant, acorn squash, macaw, bolete, fiddler crab, mobile home, dressing table, chimpanzee, jack o' lantern, toast, nipple, entlebucher, groom, sarong, cauliflower, apiary, english foxhound, deck chair, car door, labrador retriever, wallaby, acorn, short pants, standard schnauzer, lampshade, hog, male horse, martin, loudspeaker, plum, bale, partridge, water jug, shoji, shield, american lobster, nailfile, poodle, jackfruit, heifer, whippet, mitten, eggnog, weimaraner, twin bed, english springer, dowitcher, rhesus, norwich terrier, sail, custard apple, wassail, bib, bullet, bartlett, brace, pick, carthorse, ruminant, clog, screw, burro, mountain bike, sunscreen, packet, madagascar cat, radio telescope, wild sheep, stuffed peppers, okapi, bighorn, grizzly, jar, rambutan, mortarboard, raspberry, gar, andiron, paintbrush, running shoe, turnstile, leonberg, red wine, open face sandwich, metal screw, west highland white terrier, boxer, lorikeet, interceptor, ruddy turnstone, colobus, pan, white stork, stinkhorn, american coot, trailer truck, bride, afghan hound, motorboat, bassoon, quesadilla, goblet, llama, folding chair, spoonbill, workhorse, pimento, anemone fish, ewe, megalith, pool ball, macaque, kit fox, oryx, sleeve, plug, battery, black stork, saluki, bath towel, bee eater, baboon, dairy cattle, sleeping bag, panpipe, gemsbok, albatross, comb, snow goose, cetacean, bucket, packhorse, palm, vending machine, butternut squash, loupe, ox, celandine, appenzeller, vulture, crampon, backboard, european gallinule, parsnip, jersey, slide, guava, cardoon, scuba diver, broom, giant schnauzer, gordon setter, staffordshire bullterrier, conch, cherry, jam, salmon, matchstick, black swan, sailboat, assault rifle, thatch, hook, wild boar, ski pole, armchair, lab coat, goldfinch, guinea pig, pinwheel, water buffalo, chain, ocarina, impala, swallow, mailbox, langur, cock, hyena, marimba, hound, knot, saw, eskimo dog, pembroke, sealyham terrier, italian greyhound, shih tzu, scotch terrier, yawl, lighter, dung beetle, dugong, academic gown, blanket, timber wolf, minibus, joystick, speedboat, flagpole, honey, chessman, club sandwich, gown, crate, peg, aquarium, whooping crane, headboard, okra, trench coat, avocado, cayuse, large yellow lady's slipper, ski mask, dough, bassarisk, bridal gown, terrapin, yacht, saddle, redbone, shower curtain, jennet, school bus, otterhound, irish terrier, carton, abaya, window shade, wooden spoon, yurt, flat coated retriever, bull mastiff, cardigan, river boat, irish wolfhound, oxygen mask, propeller, earthstar, black footed ferret, rocking chair, beach wagon, litchi, pigeon.
\\

\subsection*{Testing Class Split}

\noindent beer, musical keyboard, maple, christmas tree, hiking equipment, bicycle helmet, goggles, tortoise, whiteboard, lantern, convenience store, lifejacket, squid, watermelon, sunflower, muffin, mixer, bronze sculpture, skyscraper, drinking straw, segway, sun hat, harbor seal, cat furniture, fedora, kitchen knife, hand dryer, tree house, earrings, power plugs and sockets, waste container, blender, briefcase, street light, shotgun, sports uniform, wood burning stove, billboard, vehicle registration plate, ceiling fan, cassette deck, table tennis racket, bidet, pumpkin, tablet computer, rhinoceros, cheese, jacuzzi, door handle, swimming pool, rays and skates, chopsticks, oyster, office building, ratchet, salt and pepper shakers, juice, bowling equipment, skull, nightstand, light bulb, high heels, picnic basket, platter, cantaloupe, croissant, dinosaur, adhesive tape, mechanical fan, winter melon, egg, beehive, lily, cake stand, treadmill, kitchen \& dining room table, headphones, wine rack, harpsichord, corded phone, snowman, jet ski, fireplace, spice rack, coconut, coffeemaker, seahorse, tiara, light switch, serving tray, bathroom cabinet, slow cooker, jalapeno, cartwheel, laelia, cattleya, bran muffin, caribou, buskin, turban, chalk, cider vinegar, bannock, persimmon, wing tip, shin guard, baby shoe, euphonium, popover, pulley, walking shoe, fancy dress, clam, mozzarella, peccary, spinning rod, khimar, soap dish, hot air balloon, windmill, manometer, gnu, earphone, double hung window, conserve, claymore, scone, bouquet, ski boot, welsh poppy, puffball, sambuca, truffle, calla lily, hard hat, elephant seal, peanut, hind, jelly fungus, pirogi, recycling bin, in line skate, bialy, shelf bracket, bowling shoe, ferris wheel, stanhopea, cowrie, adjustable wrench, date bread, o ring, caryatid, leaf spring, french bread, sergeant major, daiquiri, sweet roll, polypore, face veil, support hose, chinese lantern, triangle, mulberry, quick bread, optical disk, egg yolk, shallot, strawflower, cue, blue columbine, silo, mascara, cherry tomato, box wrench, flipper, bathrobe, gill fungus, blackboard, thumbtack, longhorn, pacific walrus, streptocarpus, addax, fly orchid, blackberry, kob, car tire, sassaby, fishing rod, baguet, trowel, cornbread, disa, tuning fork, virginia spring beauty, samosa, chigetai, blue poppy, scimitar, shirt button.

\begin{figure*}[!ht]
\centering
\includegraphics[width=1.0\linewidth]{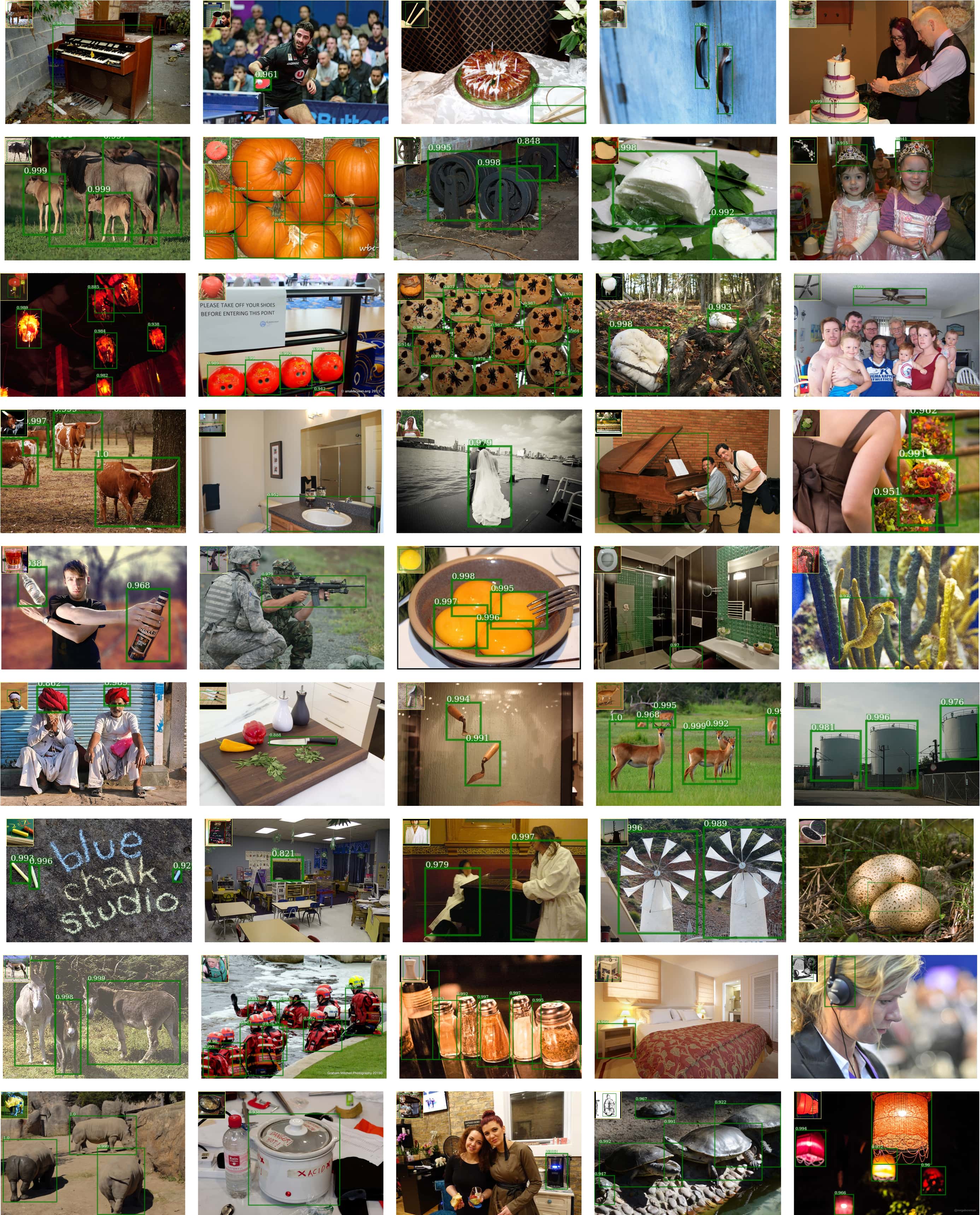}
\caption{Qualitative 1-shot object detection results on our test set.  We visualize the bounding boxes with score larger than 0.8.}
\label{fig:vis_1}
\vspace{0.2cm}
\end{figure*}

\begin{figure*}[!ht]
\centering
\includegraphics[width=1.0\linewidth]{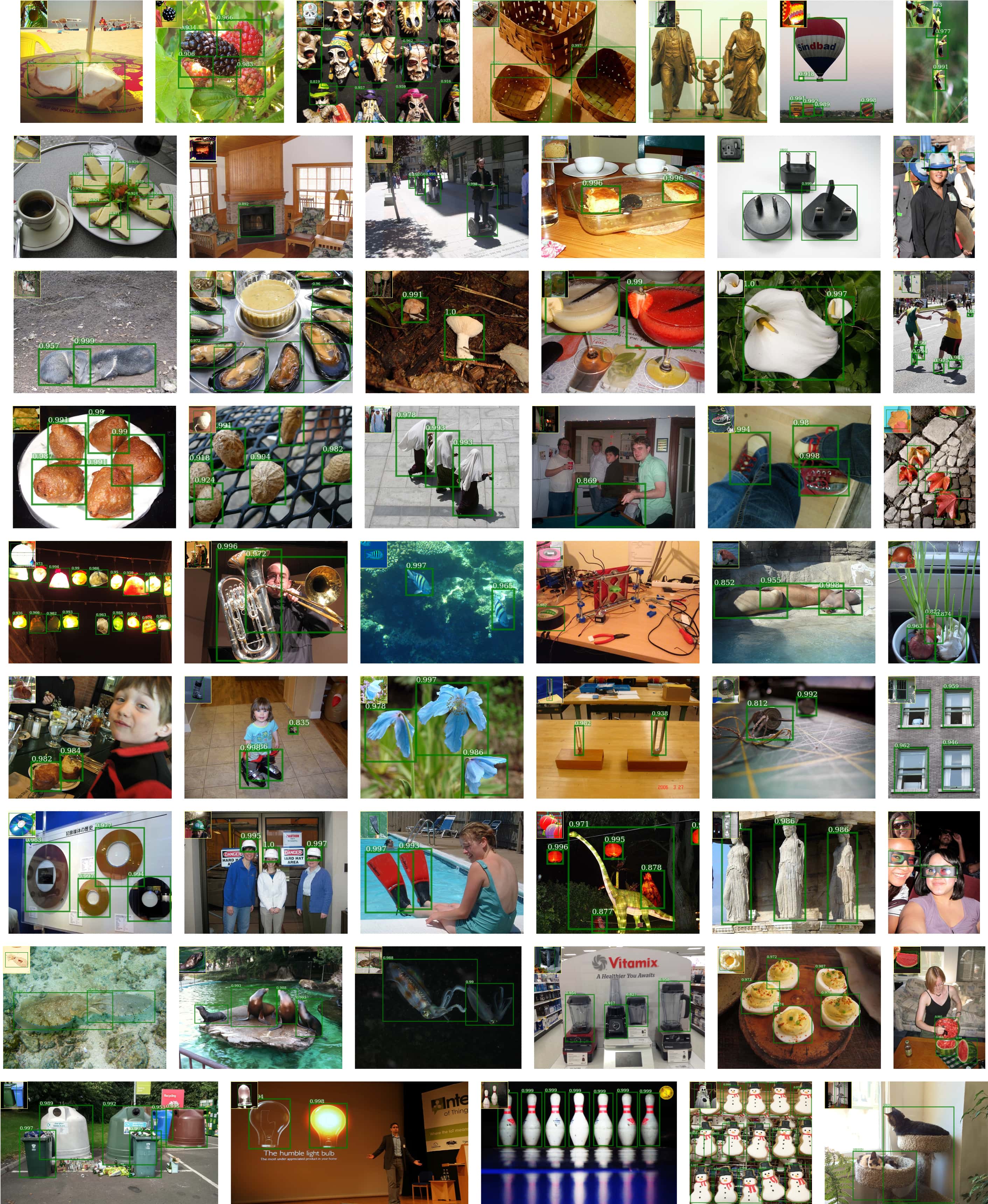}
\caption{Qualitative results of our 1-shot object detection on test set. We visualize the bounding boxes with score larger than 0.8.}
\label{fig:vis_2}
\vspace{0.4cm}
\end{figure*}

\begin{figure*}[!ht]
\centering
\includegraphics[width=0.99\linewidth]{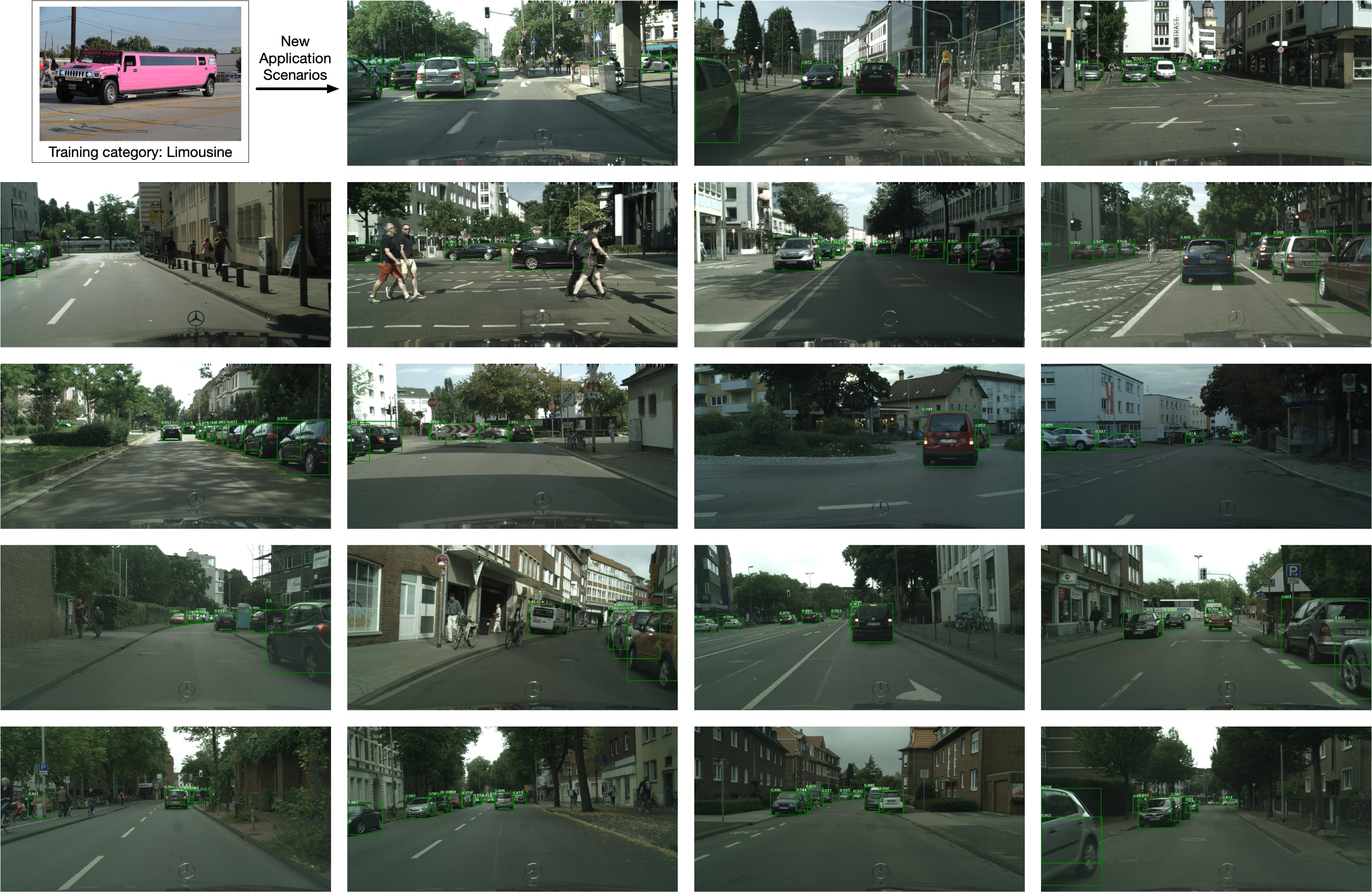}
\caption{Qualitative results of our 5-shot car detection on Cityscapes. We visualize the bounding boxes with score larger than 0.8. The first image is a training example.}
\label{fig:cityscapes}
\end{figure*}

\begin{figure*}[!ht]
\centering
\includegraphics[width=0.99\linewidth]{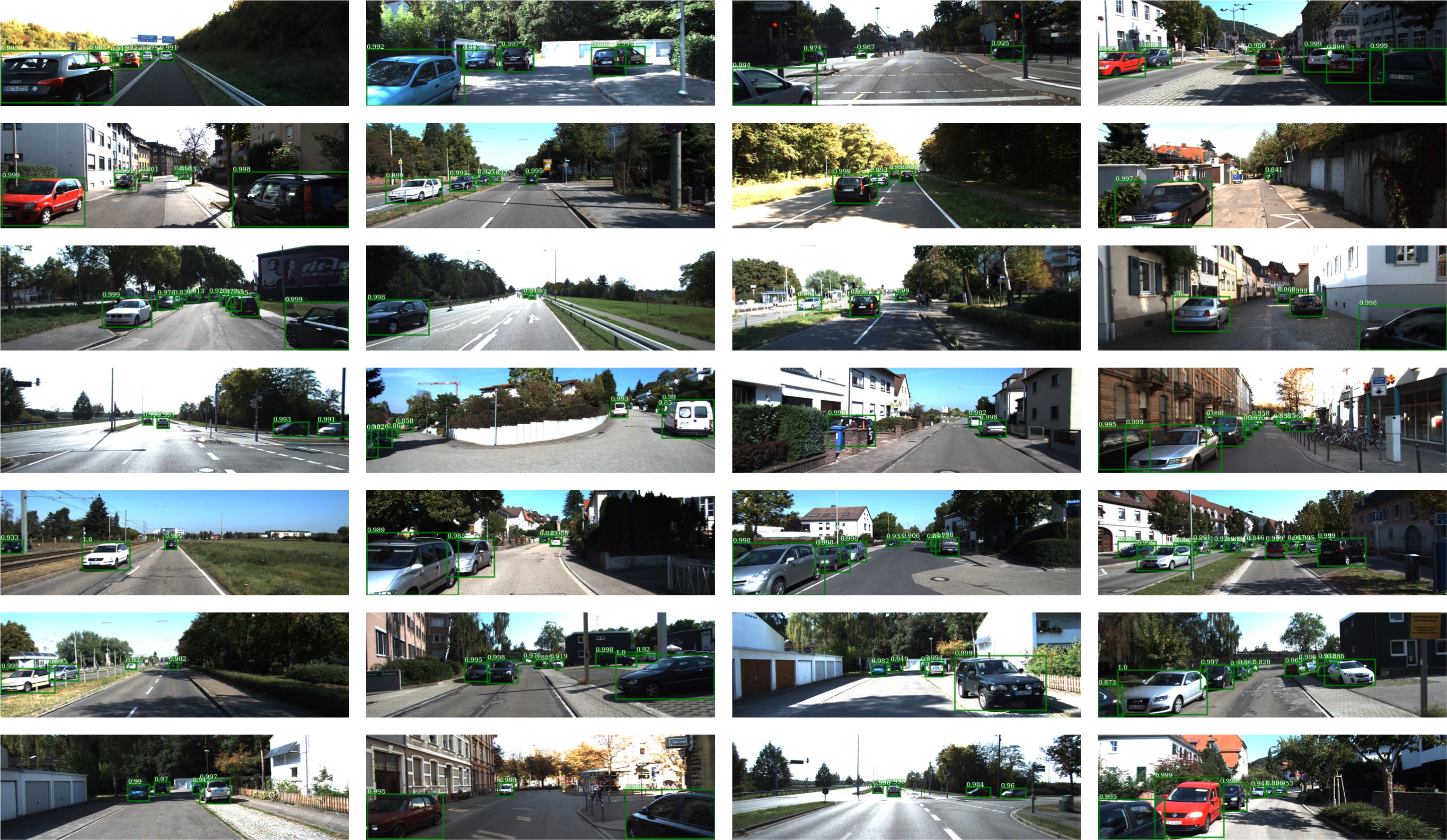}
\caption{Qualitative results of our 5-shot car detection on KITTI. We visualize the bounding boxes with score larger than 0.8.}
\label{fig:kitti}
\end{figure*}

{\small
\bibliography{main}
}

\end{document}